\documentclass[conference]{IEEEtran}
\IEEEoverridecommandlockouts
\usepackage{cite}
\usepackage{amsmath,amssymb,amsfonts}
\usepackage{algorithmic}
\usepackage{graphicx}
\usepackage{textcomp}
\usepackage{xcolor}

\usepackage{booktabs}
\usepackage{tabularx}
\usepackage{array}

\usepackage{color}

\usepackage[caption=false,listofformat=subsimple,labelformat=simple]{subfig}

\usepackage[justification=centering]{caption}
\usepackage{tikz}

\def\BibTeX{{\rm B\kern-.05em{\sc i\kern-.025em b}\kern-.08em
    T\kern-.1667em\lower.7ex\hbox{E}\kern-.125emX}}
\begin{document}

\title{SonarVoxNet: Diver Detection in 3D Bounding Box using 3D Sonar\\
}

\author{\IEEEauthorblockN{
Eugene Park, Jiwon Lee, Seyoung Kan, Trung Dong, Xiaomin Lin, and Jane Shin}
}


\maketitle

\begin{abstract}
Autonomous underwater vehicles (AUVs) assisting human divers must continuously track not only the diver's 3D position but also their full-body orientation. However, vision-based perception is unreliable underwater, and forward-looking sonar -- despite being widely used -- discards the elevation information needed for orientation estimation, posing a fundamental limitation. Recently commercialized 3D sonar preserves elevation but produces sparse, noisy returns, and existing detectors are built for dense LiDAR data and for targets that remain upright and rotate only about the yaw axis (e.g., vehicles, pedestrians), making them unable to represent a freely pitching and rolling diver. To address this gap, we present two contributions. First, SonarVoxNet adapts a voxel-based encoder and an anchor-free center-based detection head to 3D sonar data, replacing the conventional yaw-only rotation representation with a continuous 6D rotation parameterization to predict full 9-DoF oriented bounding boxes—to our knowledge, the first 3D sonar diver detector to do so. Second, Diver3D is the first public 3D sonar dataset with full 3D orientation labels for divers in diverse, non-upright poses, collected at a natural cave-diving site. Through controlled ablations over the backbone and detection head, we show that the dominant factor behind accurate 3D sonar-based diver detection is the transition from yaw-only rotation to full-SO(3) rotation. This transition substantially improves detection accuracy and reduces orientation error. These results demonstrate that full-body diver orientation is recoverable from 3D sonar alone, laying the groundwork for future work on diver pose estimation and diver-robot interaction.
\end{abstract}


    
    


\section{Introduction}
Autonomous underwater vehicles (AUVs) are increasingly deployed to assist marine operations, from infrastructure inspection to scientific exploration; however, many tasks still require a human diver in the loop. For an AUV to safely and usefully assist such a diver, it must continuously track the diver's position and 3D body orientation -- a prerequisite for any context-aware human-robot interaction underwater. However, vision-based perception \cite{WangYolov1024, ZhaoRTDETR24}, despite its success on land, is poorly suited to this task: light attenuation, scattering, and turbidity frequently render optical cameras unreliable underwater. The community has therefore turned to acoustic imaging, and in particular forward-looking sonar (FLS), which has been extensively used for detecting objects on the seabed.

\begin{figure}[t]
    \centering
    \includegraphics[width=0.9\linewidth]{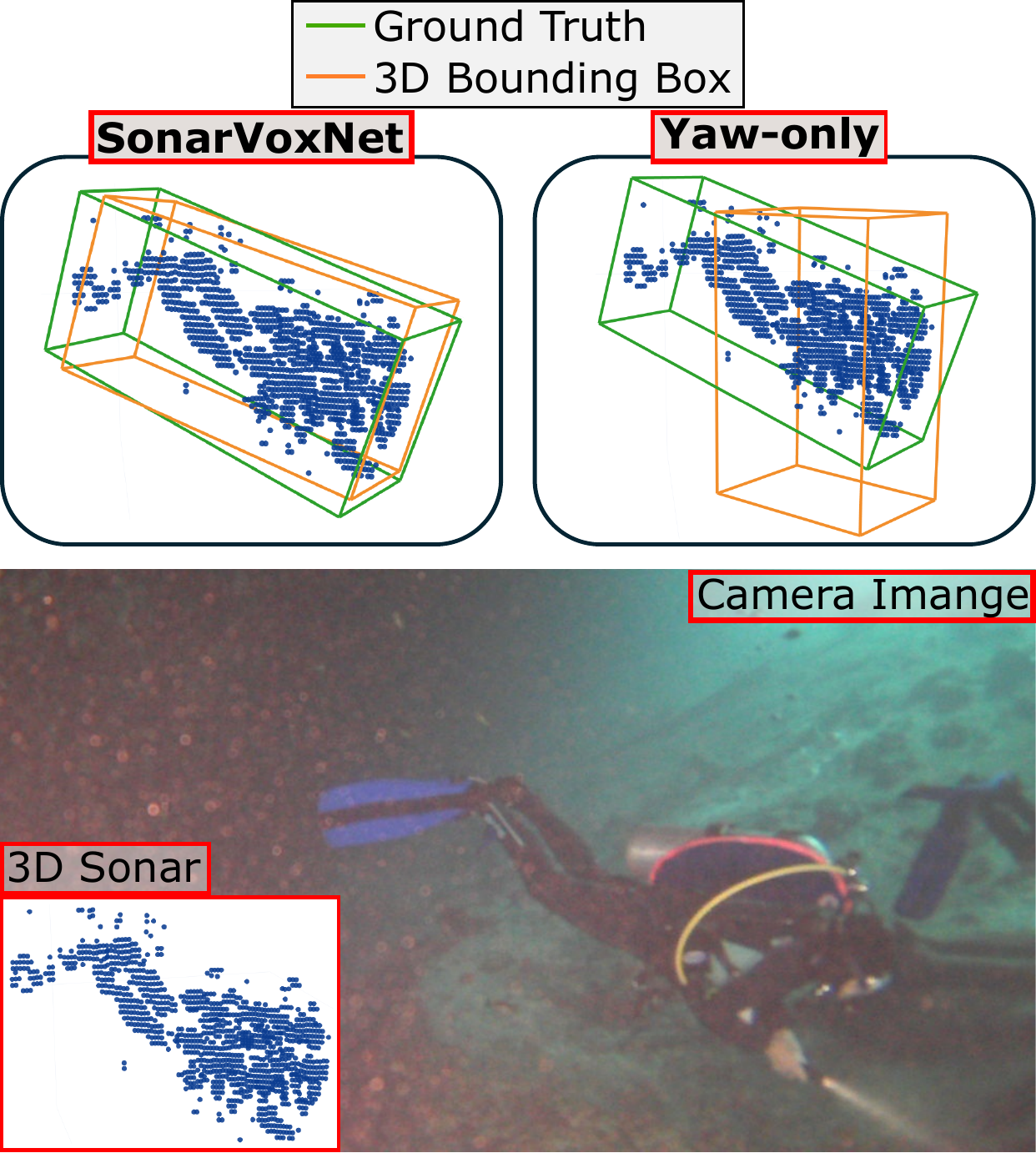}
    \caption{3D Bounding Box Prediction for Underwater Diver Detection from SonarVoxNet using 3D Sonar Sensor}
    \label{fig:main_fig}
\end{figure}

Applying FLS to diver monitoring, however, exposes a fundamental geometric bottleneck: FLS projects 3D spatial structure onto a single 2D acoustic plane, discarding elevation. This loss is often tolerable for rigid, seabed-bound objects, but not for a diver, whose 3D orientation in the water column is precisely the quantity of interest.

This limitation motivates volumetric sensing, recently made practical by the commercialization of 3D sonar sensors, which preserve elevation alongside range and azimuth. The transition from 2D to 3D acoustic data\cite{penumarti2026calibration}, however, trades one bottleneck for several new ones. Acoustic returns from a non-rigid human body are sparse, non-uniform, and corrupted by multipath noise, causing architectures built for dense LiDAR and RGB data\cite{dong2026post,liu2023bevfusion} to fail. Just as importantly, the detection heads built for such dense sensors assume that targets are upright and rotate only about the vertical (yaw) axis. This assumption is violated outright by a free-swimming diver, who pitches and rolls continuously; a yaw-only detector can localize a diver while completely misrepresenting the out-of-plane orientation that HRI most depends on.

We note that this paper targets diver detection -- an oriented 3D bounding box capturing position, size, and full-body orientation -- rather than per-joint skeletal pose estimation, which we view as a subsequent problem that first requires solving accurate 3D-oriented detection. To the best of our knowledge, there is no existing 3D sonar method that recovers even the coarse body orientation of a diver using real 3D sonar, and there is no existing 3D sonar dataset that captures divers in sufficiently diverse, non-upright poses to train or evaluate one.

To address the aforementioned gaps:
\begin{itemize}
\item We present \emph{SonarVoxNet}, which adapts a VoxelNet \cite{zhou2017voxelnetendtoendlearningpoint} volumetric encoder and a CenterPoint \cite{yin2021center} anchor-free head to sparse, noisy 3D sonar returns, replacing yaw-only output with a continuous 6D rotation parameterization \cite{zhou2020continuityrotationrepresentationsneural}. To our knowledge, this is the first 9-DoF diver detector for 3D sonar.
\item We present Diver3D, comprising 2 hours, 37,010 frames, and 34,568 annotated instances across 36 scenes, collected at Blue Grotto, a natural spring cave-diving site near Gainesville, FL, where divers exhibit realistic pitch and roll rarely seen in pool settings. To our knowledge, this is the first public 3D sonar dataset with full 3D orientation labels for human divers.
\item Through a controlled backbone-and-head ablation, we show that adopting full-SO(3) rotation, rather than yaw-only, is the dominant factor behind accurate 3D detection on sonar data, outweighing backbone choice. We further show that a sonar-specific, orientation-preserving augmentation strategy yields additional gains in detection accuracy and orientation fidelity, and we release our full evaluation protocol as an open baseline for underwater 3D perception research.
\end{itemize}

\section{Related Works}

Underwater object detection has been studied extensively. Camera-based perception \cite{IslamUnderstandingHumanMotion18, IslamDiverFollowingAlgorithm19} is useful but becomes unreliable under light attenuation, scattering, and turbidity, motivating a shift toward acoustic sensing. Forward-looking sonar (FLS) has been used for underwater object detection \cite{Kvasić2DFLSDiverDetection19}, but its projection of 3D space onto a single 2D acoustic plane makes it challenging to recover a diver's 3D body orientation. 3D sonar-based detection has recently emerged to address this projection loss, enabled by 3D sonar sensors that have supported underwater tasks such as dense mapping \cite{burgul2025underwater} and motion planning \cite{vadakkekuruppath2025shrums}. SonarPoint \cite{ZhaoSonarPoint25} is the only prior work in this direction, and it targets generic underwater objects rather than divers. This paper is the first to address diver detection with full 3D orientation using 3D sonar, and we introduce Diver3D, the first public 3D sonar dataset with full 3D orientation labels for human divers, to support this task.

While diver detection from 3D sonar is unexplored, human detection from point-cloud-like sensors has been studied extensively in the context of autonomous driving. Backbones for this task are broadly point-based \cite{QiPointNet17, QiPointNet++17, ShiPointRCNN19, QiVoteNet19} or voxel-based \cite{zhou2017voxelnetendtoendlearningpoint, YanSECOND18, yin2021center, chen2023voxelnextfullysparsevoxelnet, LangPointPillars19}. Point-based methods operate directly on raw points but scale poorly with point count, making them less suitable for real-time use. Voxel-based methods instead discretize the point cloud into a regular grid, enabling efficient convolutional processing at comparable accuracy. We adopt a voxel-based backbone for this reason, following the encoding scheme of VoxelNet \cite{zhou2017voxelnetendtoendlearningpoint}.

On top of a backbone, an architecture attaches a detection head, which is either anchor-based \cite{zhou2017voxelnetendtoendlearningpoint, YanSECOND18, LangPointPillars19} or anchor-free \cite{yin2021center, chen2023voxelnextfullysparsevoxelnet}. Anchor-based heads match densely predefined boxes to ground truth via Intersection over Union (IoU), requiring careful tuning of anchor priors and Non-Maximum Suppression (NMS) post-processing \cite{NeubeckNMS06} to remove duplicates. However, this is sensitive to hyper-parameters and requires massive computation time. Anchor-free, center-based heads instead predict each object as a local peak in a heatmap, avoiding both issues. We adopt this anchor-free formulation, as it better matches the diver-detection regime: divers are sparse (one to two per frame) and vary continuously in orientation, for which fixed anchor priors are a poor fit. Table \ref{tab:related_work_comparison} summarizes the backbone and head design of the five representative methods we compare against in our experiments.

\begin{table}
\centering
\caption{Architecture comparison to existing related works.}
\label{tab:related_work_comparison}
\scriptsize
\begin{tabular}{@{}lcccc@{}}
\toprule
Method & Backbone & Head & Rotation \\
\midrule
VoxelNet~\cite{zhou2017voxelnetendtoendlearningpoint}
    & Dense voxel & Anchor-based & Yaw only \\
SECOND~\cite{YanSECOND18}
    & Sparse voxel & Anchor-based & Yaw only \\
PointPillars~\cite{LangPointPillars19}
    & Pillar (2D voxel) & Anchor-based & Yaw only \\
CenterPoint~\cite{yin2021center}
    & Sparse voxel & Anchor-free (center) & Yaw only \\
VoxelNeXt~\cite{chen2023voxelnextfullysparsevoxelnet}
    & Fully sparse voxel & Anchor-free (center) & Yaw only \\
\textbf{SonarVoxNet (ours)}
    & \textbf{Sparse voxel} & \textbf{Anchor-free (center)} & \textbf{Full-SO(3)} \\
\bottomrule
\end{tabular}
\end{table}

All of these methods, however, have been developed for autonomous-driving benchmarks \cite{GeigerKITTI13}, where pedestrians and vehicles remain upright and rotate only about the vertical (yaw) axis; their detection heads parameterize orientation with a single yaw angle accordingly. This assumption breaks down underwater: a free-swimming diver pitches and rolls continuously, and a yaw-only box can localize a diver while completely misrepresenting their out-of-plane orientation. None of the five baselines in Table \ref{tab:related_work_comparison} can represent this by construction. Addressing this gap -- predicting a full, continuous 3D rotation based on \cite{zhou2020continuityrotationrepresentationsneural} rather than a single yaw angle -- is the architectural contribution of SonarVoxNet, detailed in Section \ref{sec:sonarvoxnet}.

\section{SonarVoxNet}
\label{sec:sonarvoxnet}

\subsection{Overall Architecture}
\label{sec:A}

\begin{figure*}[t]
      \centering
      \includegraphics[width=\textwidth]{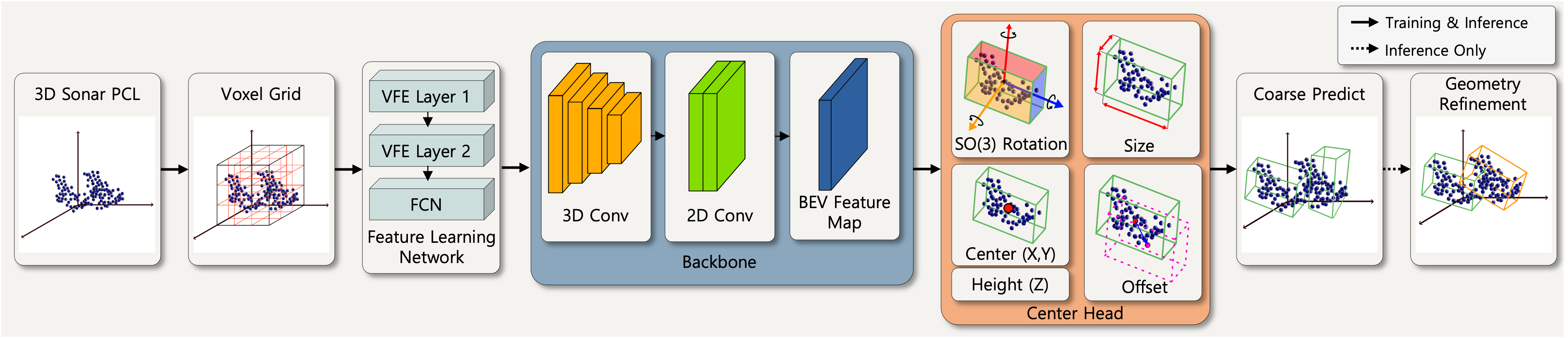}
      \caption{The architecture of the presented SonarVoxNet}
    \label{fig:arch}
\end{figure*}



We present SonarVoxNet, a voxel-based 3D diver detector adapted from previous LiDAR detection pipelines to the diver detection task using 3D sonar data. SonarVoxNet follows a `voxelize-encode-detect' pipeline. Given a 3D sonar point cloud $\mathcal{P}=\{\mathbf{p}_i\}$, where each point $\mathbf{p}_i=(x_i,y_i,z_i,s_i)$ carries a Cartesian coordinate and a scalar return feature $s_i$, we quantize the sensor field-of-view into a regular voxel grid of resolution $(v_x,v_y,v_z)$. Points falling in the same voxel are aggregated by a \emph{Voxel Feature Encoder} (VFE) into a fixed-length descriptor, following VoxelNet \cite{zhou2017voxelnetendtoendlearningpoint}. The resulting dense voxel tensor is processed by a 3D convolutional middle stage that gradually collapses the z-axis, producing a dense 2D feature map $\mathbf{F}\in\mathbb{R}^{C\times H\times W}$. A 2D convolutional encoder then refines $\mathbf{F}$ into the Bird-eye-view (BEV) feature, the shared representation consumed by the detection head.

On this BEV representation, we attach a \emph{center-based} detection head (Section~\ref{sec:B}) rather than an anchor-based one. This choice is motivated by the diver-detection regime: targets are sparse (on the order of one to two divers per frame) and vary continuously in orientation, for which anchor priors are a poor fit; a center head predicts each object as a local peak in a BEV heatmap and regresses its geometry directly.

Finally, the regressed geometry accompanied with location, forms 3D bounding boxes which go through causal geometry refinement~\ref{sec:refine}. This, inference-time causal module focuses on the out-of-BEV-plane geometry, fitting the box more to the divers freely hovering underwater.

\subsection{Detection Head}
\label{sec:B}

The head operates on the shared BEV feature map and consists of several sibling branches, each a small convolutional sub-network.

\noindent\textbf{Center localization:}
A heatmap branch predicts a single-diver class score map $\hat{\mathbf{H}}\in[0,1]^{H\times W}$, whose local maxima correspond to detected object centers projected onto the BEV plane. Because BEV quantization discards sub-pixel position, an offset branch regresses the continuous displacement $(\delta_x, \delta_y)$ from the discretized peak to the true center, and a separate branch regresses the object height $z$. Together these recover the full 3D center $\mathbf{c}=(x,y,z)$.

\noindent\textbf{Size.}
A dimension branch regresses the box extent $(l,w,h)$ in log space to stabilize the regression across the diver size distribution.

\noindent\textbf{Rotation.}
Standard BEV detectors parameterize orientation by a single yaw angle, implicitly assuming upright targets---valid for on-road vehicles or pedestrians but not for divers, who pitch and roll freely in the water column. We therefore predict a full 3D rotation $\mathbf{R}\in SO(3)$. To make this amenable to gradient-based learning we adopt the continuous 6D rotation representation of Zhou et al.
: the network outputs two vectors $(\mathbf{a}_1,\mathbf{a}_2)\in\mathbb{R}^{6}$, which are mapped to a rotation matrix by Gram--Schmidt orthonormalization,
\begin{equation}
    \label{eq:gs}
    \mathbf{b}_1=\frac{\mathbf{a}_1}{\lVert\mathbf{a}_1\rVert},\qquad\mathbf{b}_2=\frac{\mathbf{a}_2-(\mathbf{b}_1^{\!\top}\mathbf{a}_2)\,\mathbf{b}_1}
    {\lVert\mathbf{a}_2-(\mathbf{b}_1^{\!\top}\mathbf{a}_2)\,\mathbf{b}_1\rVert},\qquad\mathbf{b}_3=\mathbf{b}_1\times\mathbf{b}_2,
\end{equation}
with $\mathbf{R}=[\,\mathbf{b}_1\;\mathbf{b}_2\;\mathbf{b}_3\,]$.
Unlike Euler-angle or quaternion outputs, the 6D representation is continuous and singularity-free over all of $SO(3)$, which we find essential for regressing the wide range of diver poses in the data.

A detected object is thus described by the tuple
$(\mathbf{c},l,w,h,\mathbf{R})$, a full 9-DoF oriented 3D bounding box.

\subsection{Training Objective}
\label{sec:C}

SonarVoxNet is trained end-to-end with a multi-task loss combining center classification and box regression:
\begin{equation}
    \label{eq:total}
    \mathcal{L}=\mathbf{L}_{\mathrm{hm}}+\lambda_{\mathrm{reg}}\,\mathcal{L}_{\mathrm{reg}}.
\end{equation}

\noindent\textbf{Heatmap loss}
The ground-truth heatmap $\mathbf{H}$ is rendered by splatting each object center with a 2D Gaussian whose radius adapts to the object size. We supervise $\hat{\mathbf{H}}$ with a penalty-reduced Gaussian focal loss,
\begin{equation}
    \label{eq:hm}
    \mathcal{L}_{\mathrm{hm}}=-\frac{1}{N}\sum_{u,v}
    \begin{cases}
        \bigl(1-\hat{H}_{uv}\bigr)^{\alpha}\log\hat{H}_{uv},
          & H_{uv}=1,\\[6pt]
        \bigl(1-H_{uv}\bigr)^{\beta}\,\hat{H}_{uv}^{\,\alpha}\log\bigl(1-\hat{H}_{uv}\bigr),
          & \text{otherwise},
    \end{cases}
\end{equation}
where $N$ is the number of objects and $(\alpha,\beta)$ are focal hyperparameters. The $\bigl(1-H_{uv}\bigr)^{\beta}$ term down-weights penalties near a true center, tolerating the ambiguity of the Gaussian target.

\noindent\textbf{Regression loss}
At each ground-truth center we regress the sub-voxel offset $\boldsymbol{\delta}=(\delta_x,\delta_y)$, height $z$, log-dimensions $(\log l,\log w,\log h)$, and the continuous 6D rotation target. Let $R^{\star}=[\mathbf{r}^{\star}_{1}\ \mathbf{r}^{\star}_{2}\ \mathbf{r}^{\star}_{3}]$ be the ground-truth rotation matrix. Its 6D target is defined as $\boldsymbol{\rho}^{\star}=[(\mathbf{r}^{\star}_{1})^{\mathsf T},(\mathbf{r}^{\star}_{2})^{\mathsf T}]^{\mathsf T}$.
We combine all regression targets into  $\mathbf{t}^{\star}=(\delta_x^{\star},\delta_y^{\star},z^{\star},\log l^{\star},\log w^{\star},\log h^{\star},(\boldsymbol{\rho}^{\star})^{\mathsf T})^{\mathsf T}$ and apply an $\ell_{1}$ minimization
\begin{equation}
    \label{eq:reg}
    \mathcal{L}_{\mathrm{reg}}=\frac{1}{N}\sum_{u,v}M_{uv}\left\|\hat{\mathbf{t}}_{uv}-\mathbf{t}^{\star}_{uv}\right\|_{1},
\end{equation}
 where $M_{uv}$ indicates whether $(u,v)$ is a ground-truth center cell and $N=\max(1,\sum_{u,v}M_{uv})$. At inference, the predicted 6D component is mapped to a valid rotation matrix using the Gram--Schmidt orthonormalization in Eq.~\ref{eq:gs}.


\subsection{Causal Geometry Refinement}
\label{sec:refine}
The detector decodes each frame independently, which can introduce jitters in the predicted box geometry. We therefore apply a lightweight causal refinement stage at inference-time. Detections are first associated into short tracks using bird's-eye-view overlap. Within each track, the stage recursively refines the box center, rotation, and size using only observations up to the current frame. Specifically, the center is refined by an interacting multiple-model (IMM) filter in Euclidean space $\mathbb{R}^3$~\cite{Blom1988TheIM}, and the rotation by an analogous filter in the tangent space of $SO(3)$~\cite{solà2021microlietheorystate}, and the size by a conservative shrink-only temporal estimate:
\begin{equation}
    \label{eq:refine-map}
    \begin{aligned}
\hat{\mathbf{c}}_t
&=
\mathrm{IMM}_{\mathbb{R}^3}
\!\left(
\mathbf{c}_t + g R_t \boldsymbol{\delta}_{\mathbf{c},t}
\right),\\
\hat{R}_t
&=
\mathrm{IMM}_{SO(3)}
\!\left(
R_t\exp(\boldsymbol{\delta}_{\mathbf{r},t})
\right),\\
\hat{\mathbf{s}}_t
&=
\min\!\left(
\mathbf{s}_t,\bar{\mathbf{s}}_{1:t}
\right).
    \end{aligned}
\end{equation}
Here, $(\hat{\mathbf{c}}_t,\hat{\mathbf{s}}_t,\hat{R}_t)$ is the refined counterparts of the raw box $(\mathbf{c}_t,\mathbf{s}_t,R_t)$. The learned residuals $\boldsymbol{\delta}_{\mathbf{c}},\boldsymbol{\delta}_{\mathbf{r}}$ are predicted in the box-local frame. For the center, $R_t$ rotates residual into the sensor frame, and we damp the correction by $g=\tfrac12$. The center and rotation IMM filters use Kalman updates with learned measurement uncertainty and a fixed sensor quantization floor. For rotation, the uncertainty is the point-conditioned learned tangent covariance, whereas the fixed sensor voxel-quantization floor for center. For size, $\bar{\mathbf{s}}_{1:t}$ is a causal temporal estimate, restricting the update to shrink-only.  

All refinement operations depend only on frames $1{:}t$. The stage also never promotes a sub-threshold candidate or removes an existing one, consequently confining its effect to box geometry, while detection recall remains bounded by the detector. We treat it as a secondary, inference-time contribution and quantify it, together with the per-axis design choices, in Section~\ref{ssec:results-refine}.

\section{Dataset}
\subsection{Data Collection}
The dataset has been collected at Blue Grotto, an underwater cave system in Williston, Florida, across seven separate field sessions. The vehicle has captured the footage of both natural and human-altered underwater settings, including divers, fish, underwater robots, submerged platforms, pipes, and cave passages, along with dedicated sequences of calibration targets. 

\subsection{Labeling Process}
For each sonar frame, a dedicated full $SO(3)$ 3D bounding box is given to each diver on the frame. Annotators identify diver returns and then place an oriented 3D bounding box around each instance in the sonar centric coordinate frame. The box is adjusted to enclose the complete perceived body extent. When a diver is partially observed, annotators label the plausible full body extent if the partial observation could be reliably identified as a diver, instead of fitting a box only to the visible returns. This procedure provides geometrically consistent supervision for partially observed targets.

To maintain a consistent orientation convention, each box is defined with respect to a diver-centric right-handed body frame $\{\mathbf{e}_x,\mathbf{e}_y,\mathbf{e}_z\}$. The $\mathbf{e}_x$-axis points along the forward direction of the diver's torso, and $\mathbf{e}_y$ follows the shoulder-to-shoulder direction from the diver's right to left. The remaining axis is determined by the right-hand rule, $\mathbf{e}_z=\mathbf{e}_x\times\mathbf{e}_y$. Accordingly, the box orientation is represented by $\mathbf{R}=[\mathbf{e}_x~\mathbf{e}_y~\mathbf{e}_z]\in SO(3)$, which maps the diver's body frame to the sonar frame; its equivalent unit quaternion is also recorded. The final dataset comprises 36 scenes, 37,010 frames, and 34,568 annotated sonar diver instances.


\subsection{Analysis of the Dataset}
\label{sec:data-analysis}
We divide our data into train, validation, and test sets. The train set comprises 21 scenes with 20,861 frames, the validation set comprises 7 scenes with 8,037 frames, and the test set consists of 8 scenes with 8,112 frames. The dataset contains only one class, diver. Divers do not appear in every frame; however, when present, one to four divers with diverse rotations appear per frame. In terms of the number of divers, no divers appear in 27.6\% of frames, a single diver appears in 55.5\% of frames and multiple divers appear in the remaining frames. We also define `tilt' as the angle between the global z-direction and the $\mathbf{e}_{x}$-direction; the tilt angle distribution includes standing (tilt$<15^\circ$) in 34.8\% of cases, lying ($75^\circ<$tilt$<105^\circ$) in 30.6\%, and the others in the remaining 34.5\%. The number of 3D sonar points in each frame ranges from 428 to 11,965 and approximately follows a normal distribution. 
The spatial range of the points is $[0.13m, 16.02m]\times[-11.33m, 11.31m]\times[-5.48m, 5.48m]$. In detail, we set the implementation grid to $[0.0m, 12.0m]\times[-5.0m, 5.0m]\times[-2.5m, 2.5m]$ such that 89.09\% of points and 99.87\% of diver class objects are included within our grid.

\section{Experiments}
\noindent\textbf{Dataset}
We evaluate SonarVoxNet on Diver3D, a single-class 3D sonar dataset. The data are split scene-wise to prevent leakage between temporally adjacent observations. The train/validation/test partitions are conducted as Section~\ref{sec:data-analysis}. 
The released \texttt{splits.json} fixes all scene identifiers. 
Validation data are used for model checkpoint selection, while the test split is reserved for reporting.

Each input point $\mathbf{p}_i=(x_i,y_i,z_i,s_i)$ is passed through the VFE, into the sparse convolutional backbone. Bounding-box geometry also follows Section~\ref{sec:sonarvoxnet}. We crop points to $[0,12]\,\mathrm{m}\times[-5,5]\,\mathrm{m}\times[-2.5,2.5]\,\mathrm{m}$. We excluded augmentation in ablation experiments if not stated otherwise; full augmentation is specified in Section~\ref{sec:aug}; validation and test data are not augmented.

\noindent\textbf{Implementation details.}

The region of interest is voxelized at $(v_x,v_y,v_z)=(0.1,0.1,0.5)$\,m, giving a $120\times100\times10$ grid with at most 35 points per voxel and $8,000$ non-empty voxels per frame. Each point is encoded with VoxelNet-style VFE layers into a 128-channel voxel-wise feature, processed by a sparse 3D convolutional middle stage that collapses height 10 $\rightarrow$ 2, and reshaped into a 128-channel BEV feature. A 2D convolutional encoder–decoder refines this into a 768-channel, $60 \times 50$ feature map ($0.2$m stride), on which independent $1 \times 1$ convolutions predict the heatmap, offset, height, log-dimensions, and 6D rotation. Gaussian center targets use minimum overlap 0.7 with focal parameters $(\alpha, \beta) = (2, 4)$; all loss terms in Section \ref{sec:sonarvoxnet} have unit weight. At inference, we retain at most $100$ local maxima after $3 \times 3$ peak suppression, discard scores below $0.30$, and apply rotated BEV NMS at IoU $0.10$.

We train in $FP32$ for $20$ epochs with batch size $4$ on one NVIDIA A100, using AdamW (peak learning rate $10^{-3}$, weight decay $10^{-2}$, gradient clipping at norm $35$) with a two-phase cosine schedule. We report the mean and population standard deviation over two random seeds. After each epoch we evaluate on a fixed validation subset and select the checkpoint maximizing the three-epoch moving average of $\mathrm{AP_{3D}}@0.35$, excluding the first 20\% of training; these selection settings and the decoding thresholds above are fixed across all experiments. Full architectural and optimization details, along with code, are provided in the supplementary material.

\noindent\textbf{Refinement module.}
\label{sec:experiments-refine}
The residual regressors of the causal refinement stage (Section ~\ref{sec:refine}) are fit only on the validation scenes under the same fixed training policy, with no hyperparameter re-selection on the test split, and are refit on all validation scenes for the final model. Refinement runs online, in acquisition order, and is scored with the same detection protocol described above.


\section{Results}

\noindent\textbf{Evaluation Metrics.}
We assess detection quality with IoU-based average precision in 3D and in bird's-eye view (AP$_{\mathrm{3D}}$, AP$_{\mathrm{BEV}}$), following standard criteria in the 3D detection literature~\cite{zhou2017voxelnetendtoendlearningpoint,YanSECOND18}. For every matched true positive we additionally report translation (ATE), scale (ASE), and orientation errors. Orientation is quantified by the full-3D geodesic error on $SO(3)$, $\Delta\theta=\min_{S\in\mathcal{S}}\arccos\!\big((\operatorname{tr}(R_{\mathrm{pred}}^{\top}R_{\mathrm{gt}}S)-1)/2\big)$ (AOE$_{\mathrm{3D}}$), minimized over the cuboid's yaw-symmetry set $\mathcal{S}=\{I,R_z(\pi)\}$; we fold only the $z$ (yaw) $\pi$ ambiguity because a diver's up/down and front/back are distinguishable. The yaw-only error AOE is its reduction to the $z$ component, and the tilt error is the angle between the predicted and ground-truth body up-axes (the third column of $R$), which isolates the out-of-plane pitch/roll and is invariant to yaw. 
Because a yaw-only criterion is blind to the out-of-plane pose that characterizes a free-swimming diver, AOE$_{\mathrm{3D}}$ is central to our analysis. To condense these terms into a single figure of merit, we adapt the nuScenes Detection Score~\cite{caesar2020nuscenesmultimodaldatasetautonomous} to our setting as the Sonar--Diver Detection Score (SDS). With the normalized true-positive errors $\mathbb{E}=\{\mathrm{ATE}/1\,\mathrm{m},\mathrm{ASE},\mathrm{AOE}_{\mathrm{3D}}/(\pi/2)\}$, we define
\begin{equation}
    \mathrm{SDS} = \frac{1}{6}\left[3\,\mathrm{mAP}_{\mathrm{3D}} + \sum_{e\in\mathbb{E}}\left(1-\min(1,e)\right)\right].
    \label{eq:sds}
\end{equation}
Here, $\mathrm{mAP}_{\mathrm{3D}}$ averages AP$_{\mathrm{3D}}$ over the IoU thresholds $\{0.30,0.35,0.40\}$; the average is over thresholds rather than classes because Diver3D contains a single object class. The weighting assigns half of the score to detection and distributes the remaining half across localization, scale, and full-3D orientation. We omit the velocity and attribute terms of the original NDS, for which no annotations exist.

\subsection{Overall Performance}

\begin{table}[t]
    \centering
    \small
    \caption{Performance comparison of SonarVoxNet and other 3D detectors on the Diver3D test set}
    \label{tab:benchmark_comparison}
    \renewcommand{\arraystretch}{1.2}
    \begin{tabular}{cccc}
        \toprule
        Method & AP$_{\mathrm{3D}}$@0.35 & AP$_{\mathrm{3D}}$@0.50 & SDS \\
        \midrule
        VoxelNet \cite{zhou2017voxelnetendtoendlearningpoint} & 0.126 & 0.016 & 0.323 \\
        SECOND \cite{YanSECOND18} & 0.315 & 0.016 & 0.436 \\
        PointPillars \cite{LangPointPillars19} & 0.190 & 0.018 & 0.370 \\
        CenterPoint \cite{yin2021center} & 0.324 & 0.019 & 0.433 \\
        VoxelNeXt \cite{chen2023voxelnextfullysparsevoxelnet} & 0.374 & 0.019 & 0.463 \\
        SonarVoxNet(no aug) & 0.515 & 0.233 & 0.574 \\
        \midrule
        SonarVoxNet(Ours) & 0.703 & 0.359 & 0.673 \\
        \midrule
        + refinement module & 0.774 & 0.467 & 0.719 \\
        \bottomrule
    \end{tabular}
\end{table}

We compare SonarVoxNet with previous state-of-the-art (SOTA) methods that perform well in the LiDAR domain, evaluating both on the Diver3D test set. Each method is trained for 20 epochs without data augmentation following its established training procedure with two random seeds, and we evaluate the average performance of the model on the test set using the checkpoint that achieved the highest $AP_{3D}@0.35$ on the validation set. As shown in Table \ref{tab:benchmark_comparison}, SonarVoxNet without augmentation  consistently outperforms all baselines, achieving an average improvement of 0.249 in $AP_{3D}@0.35$, 0.215 in $AP_{3D}@0.50$ and 0.131 in SDS, respectively. Ultimately, the proposed method achieves SOTA performance among 3D detectors performing 3D rotation object detection on Diver3D, a sonar-domain dataset.

\subsection{Ablation Study}
\label{sec:ablation}

\begin{table*}[ht]
    \centering
    \caption{Backbone $\times$ Head ablation on the Diver3D test set without augmentation.}
    \label{tab:ablation}
    \setlength{\tabcolsep}{3.5pt}
    \begin{tabular}{llccccccccccc}
        \toprule
        & & \multicolumn{4}{c}{AP$_{\mathrm{3D}}$ $\uparrow$} & AP$_{\mathrm{BEV}}$ $\uparrow$ & TP-IoU $\uparrow$ & ATE $\downarrow$ & ASE $\downarrow$ & AOE $\downarrow$ & AOE$_{\mathrm{3D}}$ $\downarrow$ & SDS $\uparrow$ \\
        \cmidrule(lr){3-6}
        Backbone & Head & @0.30 & @0.35 & @0.40 & @0.50 & @0.35 & (3D) & (m) & & ($^\circ$) & ($^\circ$) & \\
        \midrule
                 & Anchor + yaw                 & 0.346 & 0.242 & 0.137 & 0.022 & 0.394 & 0.430 & 0.233 & \textbf{0.332} & \textbf{33.0} & 65.1 & 0.406 \\
        Dense    & Center + yaw                 & 0.549 & 0.328 & 0.139 & 0.020 & \textbf{0.680} & 0.420 & \textbf{0.214} & 0.350 & 35.1 & 70.8 & 0.444 \\
                 & Center + 6D rotation         & \textbf{0.625} & \textbf{0.538} & \textbf{0.427} & \textbf{0.187} & 0.663 & \textbf{0.492} & 0.219 & 0.347 & 37.5 & \textbf{51.3} & \textbf{0.576} \\
        \midrule
                 & Anchor + yaw                 & 0.314 & 0.218 & 0.117 & 0.016 & 0.354 & 0.431 & 0.246 & 0.336 & 34.7 & 64.8 & 0.391 \\
        Sparse   & Center + yaw                 & 0.521 & 0.319 & 0.144 & 0.016 & \textbf{0.623} & 0.422 & 0.224 & \textbf{0.328} & \textbf{31.5} & 66.8 & 0.448 \\
                 & \textbf{Center + 6D rotation (ours)} & \textbf{0.578} & \textbf{0.515} & \textbf{0.432} & \textbf{0.233} & 0.613 & \textbf{0.510} & \textbf{0.210} & 0.330 & 34.3 & \textbf{48.5} & \textbf{0.574} \\
        \bottomrule
    \end{tabular}
\end{table*}

Our detector is defined by two architectural choices -- the 3D voxel backbone and the detection head -- which we study in a single controlled matrix. The backbone is realized with either dense 3D convolutions~\cite{zhou2017voxelnetendtoendlearningpoint}, or sparse 3D convolutions~\cite{YanSECOND18,yin2021center}. The head is one of three designs: an anchor-based regressor with a yaw angle, and an anchor-free center head with a yaw angle, or our center head with a continuous 6D full-$SO(3)$ rotation~\cite{zhou2020continuityrotationrepresentationsneural}. 
Every other component is held constant for fair ablation. This section is conducted without any augmentation, which will be deliberated separately in Section~\ref{sec:aug}. Results are reported on the held-out test set as two-seed means unless stated otherwise. 
In Table~\ref{tab:ablation}, AP$_{\mathrm{3D}}$ and AP$_{BEV}$ are reported along with true positive metrics. 
SDS aggregates mAP$_{\mathrm{3D}}$ with TP geometric errors 
per Eq.~\ref{eq:sds}. Bold marks the best head within each backbone block. The sparse full-$SO(3)$ configuration (last row) is the model carried forward.

\noindent\textbf{Anchor vs.\ center head.}
Replacing the anchor head with a center head substantially improves top-down detection on both backbones, as a single-peak center target avoids the orientation-conditioned anchor assignment, which tends to falter when targets rarely align with a small set of priors. This gain, however, is confined to the ground plane, not to 3D box quality (TP-IoU, AOE$_{3D}$), since the center head still regresses only yaw orientations of boxes in BEV.

\noindent\textbf{Effect of full-$\mathbf{SO(3)}$ rotation.}
Adding the continuous 6D rotation is the single largest factor on both backbones: AP$_{\mathrm{3D}}$@0.35 rises from 0.328 to  0.538 (dense) and 0.319 to 0.515 (sparse), with the gain widening monotonically as the IoU criterion tightens, reaching a nine- to fifteen-fold improvement at IoU@0.5. AP$_{\mathrm{BEV}}$@0.35, in contrast, barely moves while TP-IoU rises substantially, pinpointing the source of the gain: top-down detection is already solved by the center head, so the residual AP$_{\mathrm{3D}}$ gap is out-of-BEV-plane geometry -- the pitch and roll a yaw-parameterized box cannot represent. The orientation metrics confirm this from the other side: yaw-only AOE is essentially unchanged across all three heads, whereas AOE$_{\mathrm{3D}}$ falls sharply from  65--71$^\circ$ to 48--51$^\circ$, locating the improvement precisely in the out-of-plane rotation that AOE omits. This shows that detectors built for the upright objects of street scenes are fundamentally mismatched to freely tilted divers: lifting the box to $SO(3)$ is essential, not incremental.

\noindent\textbf{Dense vs.\ sparse backbone.}
The anchor$\rightarrow$center$\rightarrow$6D progression is identical for both backbones, so the choice between them is a separate, secondary question. Holding the head at our full-$SO(3)$ design, dense and sparse perform within seed noise (SDS 0.576 vs. 0.574): dense leads only on the loose-IoU AP$_{\mathrm{3D}}$ that drives SDS, while sparse is level or better on strict AP$_{\mathrm{3D}}$@0.50 and on every orientation, scale, and translation term. No accuracy argument therefore favors dense, and we adopt sparse on two further grounds: it is the field's standard since SECOND~\cite{YanSECOND18} and through CenterPoint and its successors~\cite{yin2021center,chen2023voxelnextfullysparsevoxelnet}, making it a natural substrate for future receptive-field extensions, and it is $\sim$15\% faster at matched accuracy (57 vs. 48~FPS), an advantage for typical of embedded underwater platforms.

\noindent\textbf{Rotation representation.}
\label{sec:rot-ablation}
Focusing on the full-$SO(3)$ regression as the decisive factor of the gain, we next inspect whether the conclusion is sensitive to how the rotation is parameterized. We compare four targets that are classified into two theoretical classes, under the same sparse backbone with our detection head. \emph{Continuous} representations embed $SO(3)$ without singularities~\cite{zhou2020continuityrotationrepresentationsneural}: 6D representation and 9D matrix regression with an inference-time SVD projection~\cite{zhou2020continuityrotationrepresentationsneural,ChoyTrainingOrthogonalization26}. \emph{Discontinuous} ones cannot: unit quaternions double-cover $SO(3)$, and axis--angle degenerates as the rotation magnitude approaches $\pi$. In both cases the supervision target jumps across a locus of $SO(3)$, so the network must approximate a discontinuous function to fit poses near the locus.

\begin{table}[t]
    \centering
    \setlength{\tabcolsep}{4pt}
    \begin{tabular}{lccc}
         \toprule
         Representation & AP$_{\mathrm{3D}}$@0.35 & AP$_{\mathrm{3D}}$@0.5 & AOE$_{\mathrm{3D}}$\,($^\circ$) \\
         \midrule
         \multicolumn{4}{l}{\emph{Continuous}} \\
         \underline{6D (ours)} & $0.515${\scriptsize$\pm$.030} & $\mathbf{0.233}${\scriptsize$\pm$.016} & $48.5${\scriptsize$\pm$0.7} \\
         9D + SVD & $\mathbf{0.551}${\scriptsize$\pm$.011} & $0.216${\scriptsize$\pm$.007} & $50.4${\scriptsize$\pm$1.1} \\
         \midrule
         \multicolumn{4}{l}{\emph{Discontinuous}} \\
         Quaternion & $0.521${\scriptsize$\pm$.028} & $0.193${\scriptsize$\pm$.031} & $55.9${\scriptsize$\pm$1.0} \\
         Axis--angle  & $0.504${\scriptsize$\pm$.025} & $0.207${\scriptsize$\pm$.007} & $\mathbf{45.3}${\scriptsize$\pm$2.1} \\
         \bottomrule
    \end{tabular}
    \caption{Rotation representation. \\
    \underline{Underline}: adopted. No representation leads on all three metrics.}
    \label{tab:rotation-representation}
\end{table}

No representation dominates in Table~\ref{tab:rotation-representation}: 9D leads at the loose threshold, 6D at the strict AP$_{\mathrm{3D}}$@0.5, and axis--angle on AOE$_{\mathrm{3D}}$. 
Since every representation lands between $45^{\circ}$ and $56^{\circ}$
AOE$_{\mathrm{3D}}$, and that at close range the estimate is near-random regardless of target, the orientation error floor is set by what the sensor can resolve rather than by the parameterization.

We therefore adopt 6D parameterization in that it does not depend on a measurable gap: best in strict IoU@0.5 with affordable AOE$_{3D}$, which is continuous by construction, and avoids the inference-time SVD projection and matrix-values targets that 9D requires.

\subsection{Performance over Multiple Divers}
\begin{figure}
    \centering
    \includegraphics[trim={2.4cm 19.4cm 6.0cm 3.5cm}, clip, width=\linewidth]{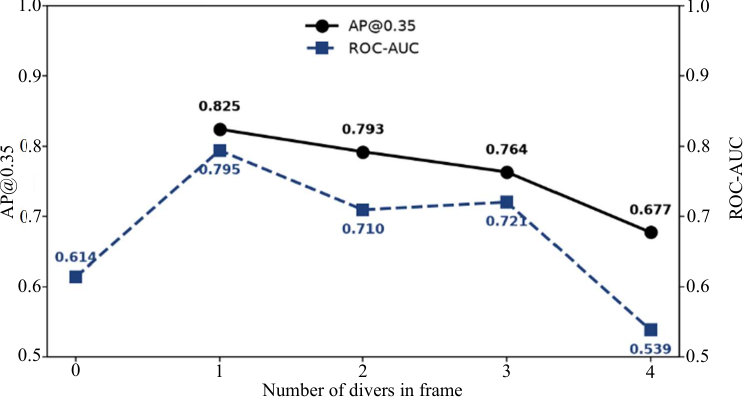}
    \caption{Diver Detection Performance per Number of Divers}
    \label{fig:DetectionPerformance}
\end{figure}

The total performance ($AP_{3D}@0.35$) of SonarVoxNet is 0.774 on the test set. When analyzing diver detection performance per number of divers, a no-diver scene structurally cannot yield $AP_{3D}@0.35$. For this reason, a Receiver Operating Characteristic (ROC) curve is used, and the Area Under the Curve (AUC) serves as the metric that captures the false-positive error. For each number of divers, the true-positive rate (TPR) is calculated across the entire test set, while the false-positive rate (FPR) is calculated using only the frames containing that number of divers. In detail, the ROC-AUC is 0.614 for no-diver frames, while $AP_{3D}@0.35$ and the ROC-AUC values for frames containing one, two, three, and four divers are shown in Fig. \ref{fig:DetectionPerformance}, respectively.

\subsection{Data Augmentation}
\label{sec:aug}
Augmentation is essential, given the noisy, class-insufficient characteristic of our underwater diver dataset. Applying augmentation raises AP$_{\mathrm{3D}}$@0.35 from $0.515$ to $0.703$ ($+37\%$ relative) with a larger relative gain at the strict IoU\,$=0.5$ threshold ($0.233\!\to\!0.359$, $+54\%$)---\emph{lifts SDS from $0.574$ to $0.673$}, (Table~\ref{tab:aug})---which is far larger effect than the few-point margins that augmentation typically buys on large-scale LiDAR benchmarks.

Compared to larger LiDAR datasets, the Diver3D sequences are highly redundant with positive-sparse nature. A quarter of the training frames are empty, and $74\%$ of the non-empty frames contain a single diver ($1.3$ divers per non-empty frame), so foreground supervision is scarce---which makes the detector vulnerable to overfitting on the redundant scene layout. Ground-truth sampling~\cite{YanSECOND18} counteracts this directly by reusing real diver returns to enrich the per-frame positive composition.

Real-Aug~\cite{zhan2023realaug} observes that plain ground-truth sampling distorts the scene crowdedness of the real data and mitigates this by annealing the number of inserted objects toward the real distribution over training. We borrow only this single idea. We adapt ground-truth sampling with object number annealing and aspect-preserving insertion constrained by the sonar view. We begin with online ground-truth sampling to a target of four divers for the first $65\%$ of training and then decrease the target by one every epoch. Unlike the zero-terminal schedule of Real-Aug, we retain one inserted target in the final stage; this specific floor prevents positive starvation on our single-class dataset and yields the best two-seed detection score. All inserted objects use aspect-preserving placement---boxes keep the bearing/body aspect relation of the observed sonar return. The curriculum with floor 1 is our best-performing setting, improving over both an all-augmented schedule (constant target four) and the zero-floor Real-Aug reference (Table~\ref{tab:aug}).

\begin{table}[t]
    \centering
    \setlength{\tabcolsep}{3pt}
    \begin{tabular}{lcccc}
         \toprule
         Training & AP$_{\mathrm{3D}}$@0.35 & AP$_{\mathrm{3D}}$@0.5 & AOE$_{\mathrm{3D}}$\,($^\circ$) & SDS \\
         \midrule
         No augmentation & $0.515$ & $0.233$ & $48.5$ & $0.574$ \\
         Constant target 4 & $0.664$ & $0.326$ & $46.5$ & $0.651$ \\
         Curriculum, zero floor & $0.676$ & $0.344$ & $\mathbf{44.6}$ & $0.662$ \\
         Curriculum, floor 1 (ours) & $\mathbf{0.703}$ & $\mathbf{0.359}$ & $46.2$ & $\mathbf{0.673}$ \\
         \bottomrule
    \end{tabular}
    \caption{Effect of augmentation on the Diver3D test set.}
    \label{tab:aug}
\end{table}

\subsection{Causal Geometry Refinement}
\label{ssec:results-refine}
Sections~\ref{sec:ablation} to \ref{sec:aug} concern the trained detector. We finally quantify the causal geometry-refinement stage of Section~\ref{sec:refine}, applied on top of the augmented final model at inference. The stage is strictly causal and adds no new detections, so its effect is confined to the geometry of boxes the detector already produced. Each paired entry in Table~\ref{tab:refine} shows one metric measured on the same box sets before versus after refinement, isolating the effects of the module alone. Values are two-seed means on the test set, and $\Delta(\%)$ is the relative chang frame Detector to +\,Refinement.

\begin{table}[t]
    \centering
    \setlength{\tabcolsep}{4pt}
    \begin{tabular}{lccc}
         \toprule
         Metric & Detector & +\,Refinement & $\Delta(\%)$ 
         \\
         \midrule
         AP$_{\mathrm{3D}}$@0.30 & $0.754$ & $\mathbf{0.808}$ & $+7.16$ \\
         AP$_{\mathrm{3D}}$@0.35 & $0.703$ & $\mathbf{0.774}$ & $+10.10$ \\
         AP$_{\mathrm{3D}}$@0.40 & $0.614$ & $\mathbf{0.709}$ & $+15.47$ \\
         AP$_{\mathrm{3D}}$@0.50 & $0.359$ & $\mathbf{0.467}$ & $+30.08$ \\
         mAP$_{\mathrm{3D}}$ & $0.690$ & $\mathbf{0.764}$ & $+10.58$ \\
         ATE (m) $\downarrow$ & $0.183$ & $\mathbf{0.169}$ & $-7.65$ \\
         ASE $\downarrow$ & $0.336$ & $\mathbf{0.327}$ & $-2.68$ \\
         AOE$_{\mathrm{3D}}$ ($^\circ$) $\downarrow$ & $46.2$ & $\mathbf{43.7}$ & $-5.41$ \\
         tilt ($^\circ$) $\downarrow$ & $21.2$ & $\mathbf{18.5}$ & $-12.74$ \\
         SDS & $0.673$ & $\mathbf{0.719}$ & $+6.84$ \\
         \bottomrule
    \end{tabular}
    \caption{Causal geometry refinement on the Diver3D test set.}
    \label{tab:refine}
\end{table}

\noindent\textbf{Effect.}
Refinement improves every geometry metrics---translation, scale, and full-3D orientation---and every AP${\mathrm{3D}}$ threshold, and the AP gain grows monotonically as the IoU criterion tightens from $+7.16\%$ at $0.30$ to $+30.08\%$ at $0.50$, lifting SDS from $0.673$ to $0.719$. This is the signature of a geometry rather than a detection effect: the module corrects boxes that are already detected, making the detected boxes fitter with better IoU, which its benefit concentrates where box quality is judged most strictly. The improvement holds on both seeds and, because the residual regressors are fit only on validation, transfers cleanly to the untouched test set. It is the same out-of-plane geometry identified in Section~\ref{sec:ablation}, reduced a second time at inference without retraining or future-frame information.

\noindent\textbf{Per-axis design.}
The three axes are not filtered identically, and the asymmetry is settled by evidence rather than preference. Each axis was taken through the same ladder on the validation split---a per-frame residual, a measurement covariance, and a causal filter---and a component was kept only if it passed a gate declared before evaluation. A learned size refiner, for example, sharpened true-positive scale but failed the gate by eroding AP$_{\mathrm{3D}}$ at the looser IoU thresholds, so size keeps its non-learned, shrink-only estimate; the center and rotation residuals passed and are retained. 

\noindent\textbf{What refinement does not do.}
The stage does not introduce new detections: it never promotes a sub-threshold candidate, so the detected-object set remains the detector's and recall is not manufactured. We verified on the validation split that the two natural ways to push past this---promoting sub-threshold candidates to recover recall, and localization-quality (IoU-aware) rescoring---do not generalize across seeds in our regime, and that the residual center error is dominated by an irreducible vertical scatter set by the sonar's range--elevation resolution rather than by a correctable bias. The remaining gap is thus a sensor- and detector-level limit, and we report causal geometry refinement as the extent of what strictly-causal post-hoc processing recovers.

\subsection{Limitations and Future Work}
\label{ssec:limitations}
First, following ground-based detector conventions, we compress the z-axis into a coarse grid before BEV projection. Future work can develop sonar domain's own decoder. Second, although Diver3D's total frame count is comparable to established LiDAR benchmarks (KITTI), its low positive density (1.3 divers per non-empty frame) leaves foreground supervision scarce. This limitation also shows the need of synthetic data augmentation. Third, full-rotation annotation for a non-rigid, freely posed diver is inherently harder to make exact, and we encountered cases where it was unclear whether the network prediction or the human label was more accurate. Finally, our overall pipeline adapts methods validated on LiDAR and adopts ground-based design choices without incorporating acoustic-sensor-specific physics or a diver-specific underwater profile analogous to ground-aware priors used on land, leaving deeper domain-specific modeling for sonar-based perception as an open direction.

\section{Conclusion}
We presented SonarVoxNet, a voxel-based 3D diver detector using a continuous 6D full-$SO(3)$ rotation head, and Diver3D, the first public 3D sonar dataset with full 3D orientation labels for divers. Our ablations show that this yaw-only-to-$SO(3)$ transition, rather than backbone or rotation-parameterization choice, is the dominant factor in accurate 3D detection. The results show that full-body diver orientation is recoverable from sparse 3D sonar alone, and we expect this to lay the groundwork for diver pose estimation and, in turn, context-aware diver-robot interaction.



\bibliographystyle{ieeetr}
\bibliography{sonarvoxnet}

\end{document}